# FUSION OF HYPERSPECTRAL AND PANCHROMATIC IMAGES USING SPECTRAL UNMIXING RESULTS

Roozbeh Rajabi [a], Hassan Ghassemian [a, *]

[a] Tarbiat Modares University, ECE Department, Tehran, Iran - (r.rajabi, ghassemi)@modares.ac.ir

**ABSTRACT:**

Hyperspectral imaging, due to providing high spectral resolution images, is one of the most important tools in the remote sensing field. Because of technological restrictions hyperspectral sensors has a limited spatial resolution. On the other hand panchromatic image has a better spatial resolution. Combining this information together can provide a better understanding of the target scene. Spectral unmixing of mixed pixels in hyperspectral images results in spectral signature and abundance fractions of endmembers but gives no information about their location in a mixed pixel. In this paper we have used spectral unmixing results of hyperspectral images and segmentation results of panchromatic image for data fusion. The proposed method has been applied on simulated data using AVRIS Indian Pines datasets. Results show that this method can effectively combine information in hyperspectral and panchromatic images.

## 1. INTRODUCTION

Remote sensing has many applications like crop classification, pollution control, resource management, etc. For these purposes, hyperspectral sensors are strong tool as they can determine chemical and physical composition of objects. Also panchromatic (PAN) sensors are the first class tools of remote sensing that have better spatial resolution than hyperspectral sensors (Borengasser et al., 2008).

Against high spectral resolution of hyperspectral images, these images have low spectral resolution. This is because of technological restrictions. To overcome this problem one approach is using PAN and hyperspectral sensor together. For example PRISMA observation system (PRISMA, 2013) consists of both hyperspectral and panchromatic sensors.

In hyperspectral images there are pixels consisting of more than one distinct material called mixed pixels. Spectral unmixing methods extract spectral signatures of these distinct materials (endmembers) and abundance fractions of them. To solve spectral unmixing problem two models can be used: linear and nonlinear. Linea mixing model (LMM) is widely used for spectral unmixing of hyperspectral images since it is simpler and in most cases there is no interaction between materials in the scene (Keshava and Mustard, 2002). Some linear unmixing methods include: VCA (Nascimento and Dias, 2005), NMF (Pauca et al., 2006), etc.

There are some works in the literature to enhance spatial resolution of hyperspectral images using spatial information available in PAN image. Gross et al. (1998) (Gross and Schott, 1998) proposed a fusion method for image sharpening using spectral mixture analysis. Another method named coupled non negative matrix factorization (CNMF) has been introduced by Yokoya et al. (2011) (Yokoya et al., 2012). Bieniarz et al. (2011) (Bieniarz et al., 2011) proposed an algorithm to fuse multispectral and hyperspectral data based on linear mixing model. A fusion method based on Indusion approach was proposed by Licciardi et al. (2012) (Licciardi et al., 2012). In this paper we have used spectral unmixing results and segmentation results of panchromatic image for data fusion. Block diagram of the proposed approach is shown in Fig. 1.

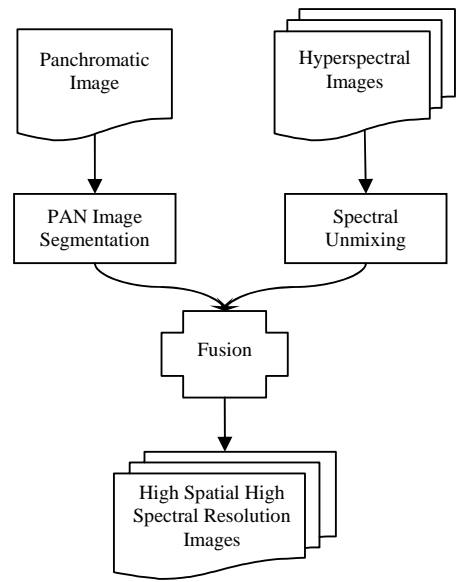

Figure 1. Block diagram of the proposed method.

The paper is organised as follows. Methodology of the proposed method including spectral unmixing, PAN image segmentation and fusion procedure is described in section 2. The process of generating simulated data is presented in section 3. Experiment on simulated data and results are presented in section 4. Finally section 5 concludes the paper.

---

* Corresponding author

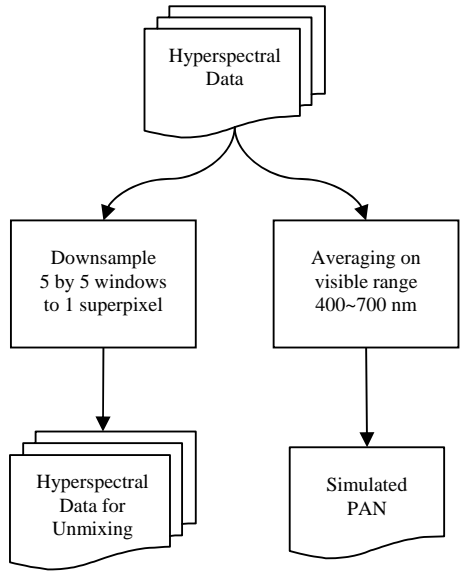

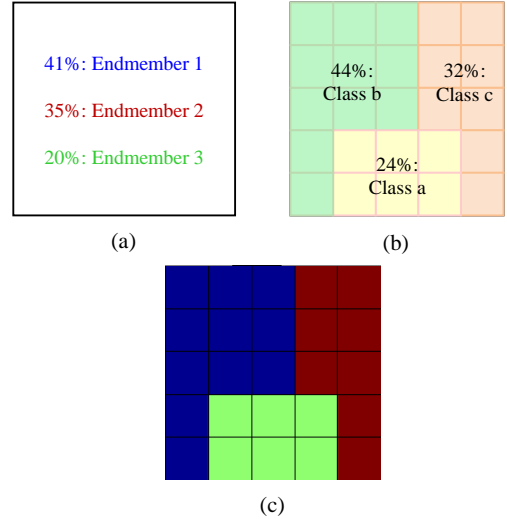

Figure 2. Procedure of generating Simulated data.

Figure 3. (a) Unmixing and (b) PAN segmentation results for one superpixel (c) Fused pixel using comparison between results.

## 2. METHODOLOGY

Block diagram of the proposed method is illustrated in Fig. 1. Firstly spectral unmixing should be done on hyperspectral data to extract spectral signature and abundance fractions of endmembers. Secondly segmentation on PAN superpixels has been done to extract spatial information. Finally information extracted from previous steps is fused together, resulting in a high spatial high spectral resolution data. Methods used for doing these steps are briefly described in the following subsections.

### 2.1 Hyperspectral Data Unmixing

Common model for spectral unmixing is linear mixing model (LMM) that many methods are based on it and is used in this work. Assume that L is the number of spectral bands. The measured spectrum (X) can be expressed by the equation (1).

$$X = SA + W \qquad (1)$$

In this equation S is spectral signature matrix of endmembers, A is abundance fraction matrix and W is an additive noise matrix. Each column of X is a linear combination of spectral signatures in S as formulated in (2).

$$x_n = Sa + w = \sum_{i=1}^{P} a_i s_i + w =$$
$$a_1 s_i + a_1 s_i + ... + a_1 s_i + w \qquad (2)$$

NMF is used for spectral unmixing of low spatial resolution hyperspectral data. For a given matrix X, NMF finds nonnegative matrix factors U and V such that:

$$X \approx UV^T \qquad (3)$$

For quantifying the quality of the approximation, cost functions based on Euclidean distance or Kullback-Leibler divergence can be used. Cost function for NMF using Euclidean distance is given in (4).

$$O = \left\| X - UV^T \right\|^2 \qquad (4)$$

Minimizing this cost function with respect to U and V subjected to $U, V \geq 0$ will lead to NMF method (Lee and Seung, 2000).

### 2.2 PAN Image Segmentation

Pixel segmenting was used previously by Aplin et al. (Aplin and Atkinson, 2001) for subpixel mapping. In this work Fuzzy C- Means is used for classification of superpixels in panchromatic image. Number of classes in each superpixel is selected equal to estimated number of endmembers in the superpixel.

### 2.3 Fusion Procedure

Comparison between abundance fractions of endmembers and pan segmentation results is used for fusion process. When abundance fractions of different materials are not distinct enough, spatial correlation (Villa et al., 2011) between current pixel and neighbour pixels has been used to resolve this ambiguity.

## 3. DATA SIMULATION

General procedure of generating simulated data for examining the proposed method is depicted in Fig. 2.

### 3.1 Hyperspectral Data

AVIRIS Indian Pines, Indiana (AVIRIS, 1992) has been used to generate simulated data. Spectral data of each class is substituted by spectral signature of materials from USGS library (Clark et al., 2007). This process results in high spatial resolution data. Image resizing (by factor 3) has been used for downsampling and generating low spatial resolution data.

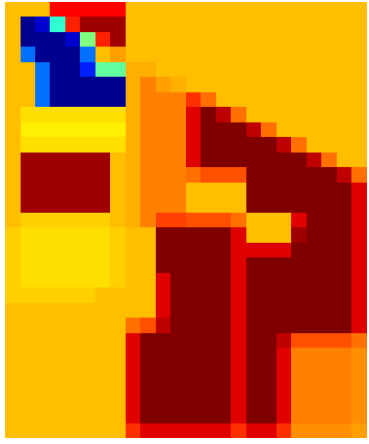

Figure 4. Band 30 of Simulated Low Spatial Resolution Data.

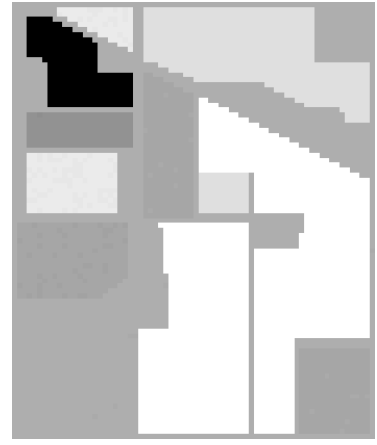

Figure 5. Simulated Panchromatic Image.

### 3.2 Panchromatic Data

PAN sensors capture image in wavelength range of 0.4 to 0.7 um. Therefore in this paper panchromatic image is simulated by resampling hyperspectral data on bands between 400 to 700 nm (Bieniarz et al., 2011).

## 4. EXPERIMENTS AND RESULTS

The proposed method has been applied to simulated data that are generated based on procedures described in section 3. Fusion result for one sample superpixel is depicted in Fig. 3. Simulated high resolution data and simulated panchromatic image are shown in Fig. 4. and Fig. 5. respectively. Original and reconstructed high spatial resolution image using proposed method is shown in Fig. 6 and Fig. 7.

### 4.1 Performance Evaluation

The performance of hyperspectral data and panchromatic image fusion method has been evaluated based on reconstruction quality of spectral signature for each single pixel in high spectral high spatial resolution image. Spectral angle error (SAE) is used for measuring the spectral reconstruction quality.

To calculate SAE first SAD measure is defined as

$$\text{SAD}_{m_i} = \cos^{-1}\left(\frac{m_i^T \hat{m}_i}{\|m_i\|\|\hat{m}_i\|}\right) \quad (5)$$

where $m_i$ is the spectral signature of $i$th pixel and $\hat{m}_i$ is its estimated value. SAE is defined as rms average value of SAD for al pixels within the scene.

For measuring spatial reconstruction quality, peak SNR (PSNR) is used. PSNR is defined as

$$\text{PSNR}_i = 10\log\left(\frac{\text{MAX}_i^2}{\text{MSE}_i}\right) \quad (6)$$

in this equation $\text{MAX}_i$ is the maximum pixel value in the $i$th band image and $\text{MSE}_i$ is defined as

$$\text{MSE}_i = \frac{1}{N}\sum_{k=1}^{N}(X - SA)^2_{i,k} \quad (7)$$

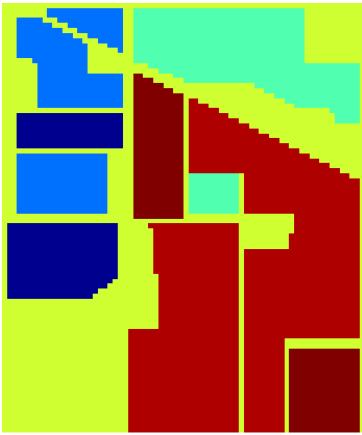

Figure 6. Band 30 of high spatial resolution simulated data.

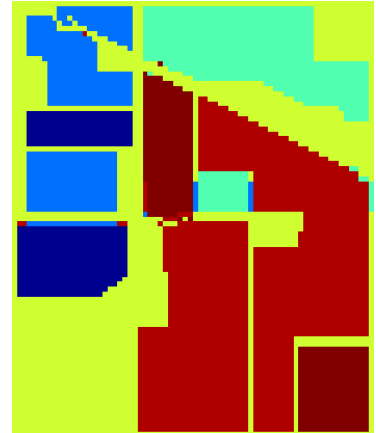

Figure 7. Band 30 of reconstructed data using prposed method.

(i,k) denotes the $k$th pixel in the $i$th band.

Smaller SAE and higher PSNR indicated higher quality of spectral and spatial reconstruction respectively (Yokoya et al., 2012).

### 4.2 Experiments

Proposed method has been applied to AVIRIS Indian Pines simulated dataset. Results based on PNSR and SAE are summarized in Table 1.

|  | PSNR (dB) | SAE (degree) |
|---|---|---|
| AVIRIS Indian Pines | 31.5 | 0.85 |

Table 1. Performance Evaluation on AVIRIS Dataset

### 5. CONCLUSION

Hyperspectral images have high spectral resolution but low spatial resolution. On the other hand panchromatic image has high spatial resolution. Hyperspectral images because of low spatial resolution contain mixed pixels. Spectral unmixing methods decompose a mixed pixel into spectral signature and abundance fraction of endmembers present in the mixed pixel but providing no information on their location. In this paper hyperspectral unmixing results and panchromatic segmentation results are fused together to obtain high spectral, high spatial resolution image. The proposed method has been applied on simulated data based on AVIRIS sensors. Results show that the proposed method can effectively combine information in hyperspectral and panchromatic images. Future work includes implementation of the proposed method on datasets containing both real hyperspectral and PAN images from one scene, using more efficient segmentation method for applying on panchromatic image and using more automated fusion method.